\documentclass[10pt,twocolumn,letterpaper]{article}

\usepackage{wacv}
\usepackage{times}
\usepackage{epsfig}
\usepackage{graphicx}
\usepackage{amsmath,amssymb} % define this before the line numbering.
\usepackage{color}
\usepackage{epsfig}
\usepackage{graphicx}
\usepackage[]{algorithm2e}
\usepackage{tabularx}
\usepackage{booktabs}
\usepackage{multirow}

\usepackage{enumitem}
\setlist{nosep}
\setlist{nolistsep}

\newcommand{\fakesubsection}[1]{\vspace{.05cm}\noindent\textbf{#1}}

\usepackage{xcolor}

\makeatletter
\DeclareRobustCommand\onedot{\futurelet\@let@token\@onedot}
\def\@onedot{\ifx\@let@token.\else.\null\fi\xspace}
\def\eg{\emph{e.g}\onedot} 
\def\ie{\emph{i.e}\onedot} 
\def\cf{\emph{cf}\onedot}

\def\etal{\emph{et al}\onedot}

\usepackage{footmisc}
\DefineFNsymbols{mySymbols}{{\ensuremath\dagger}{\ensuremath\ddagger}\S\P
   *{**}{\ensuremath{\dagger\dagger}}{\ensuremath{\ddagger\ddagger}}}
\setfnsymbol{mySymbols}

\let\tablesmall\footnotesize

\def\wacvPaperID{****} % Enter the WACV Paper ID here

\wacvfinalcopy % *** Uncomment this line for the final submission

\ifwacvfinal
\def\assignedStartPage{1} % *** Enter the assigned starting page number (instead of 9876)
\fi

\usepackage[colorlinks=true]{hyperref}

\ifwacvfinal
\else
\fi

\begin{document}

%%%%%%%%% TITLE
\title{Audio- and Gaze-driven Facial Animation of Codec Avatars}

\author{
Alexander Richard*$^1$, Colin Lea*$^{1}$\thanks{author was affiliated with Facebook at time of paper writing \mbox{\quad\ * indicates equal contribution}}, Shugao Ma$^1$, Juergen Gall$^2$, Fernando de la Torre$^1$, Yaser Sheikh$^{1}$\\
$^1$Facebook Reality Labs \quad $^2$University of Bonn \\
\footnotesize \texttt{\{richardalex, shugao, ftorre, yasers\}@fb.com, colincsl@gmail.com, gall@iai.uni-bonn.de}
}

\maketitle
%\thispagestyle{empty}

%%%%%%%%% ABSTRACT
\begin{abstract}
Codec Avatars are a recent class of learned, photorealistic face models that accurately represent the geometry and texture of a person in 3D (i.e., for virtual reality), and are almost indistinguishable from video~\cite{lombardi2018deep}. 
In this paper we describe the first approach to animate these parametric models in real-time which could be deployed on commodity virtual reality hardware using audio and/or eye tracking. 
Our goal is to display expressive conversations between individuals that exhibit important social signals such as laughter and excitement solely from latent cues in our lossy input signals. 
To this end we collected over 5 hours of high frame rate 3D face scans across three participants including traditional neutral speech as well as expressive and conversational speech. 
We investigate a multimodal fusion approach that dynamically identifies which sensor encoding should animate which parts of the face at any time.
See the supplemental video which demonstrates our ability to generate full face motion far beyond the typically neutral lip articulations seen in competing work:
\href{https://research.fb.com/videos/audio-and-gaze-driven-facial-animation-of-codec-avatars/}{https://research.fb.com/videos/audio-and-gaze-driven-facial-animation-of-codec-avatars/}
\end{abstract}

%%%%%%%%% INTRODUCTION
\section{Introduction}

% In this paper we describe a multimodal approach for driving photorealistic avatars using audio and gaze direction.
% mica2018magicleap
Advances in \textit{representing} photorealistic avatars have greatly improved in recent years~\cite{meetmike2017siggraph,siren2018gdc,MagicLeap,lombardi2018deep,lombdardi2019voxel}, however, the ability to \textit{animate} these avatars in real-time for augmented or virtual reality (AR/VR) applications remains limited~\cite{wei2019siggraph,google2017headsetRemoval}.
The state of the art in driving these avatars requires a lengthy user-specific setup process~\cite{lombardi2018deep}, custom hardware configurations not amenable to commercial AR/VR, and/or a team of technical artists mapping facial motions from a single user to their own avatar~\cite{meetmike2017siggraph,siren2018gdc}.
With ideal, sensor-heavy inputs (\ie cameras pointed clearly at the face) these approaches can accurately display a user's facial expressions, but even at best expressive speech tends to be poorly represented~\cite{wei2019siggraph}.
%In this work we investigate an approach for driving one recent photorealistic model~\cite{lombardi2018deep}, only using sensors that can be obtained with commodity hardware.
In this work we investigate an approach for driving a photorealistic model only using sensors that can be obtained with commodity hardware.
Microphones are available on all VR headsets and eye tracking is available on several developer-focused AR/VR devices.\footnote{e.g.,  \href{https://www.vive.com/eu/product/vive-pro}{Vive Pro},
\href{https://www.magicleap.com/}{Magic Leap One}, \&
\href{https://developer.qualcomm.com/hardware/snapdragon-845-vr-development-kit}{Qualcomm VR Dev Kit}
% \href{https://vr.tobii.com}{Tobii VR} \&  
} 
%  that is capable of generating plausibly accurate speech motions.
% that can be easily integrated into existing VR headsets. 

\begin{figure}
    \centering
    \includegraphics[scale=0.3]{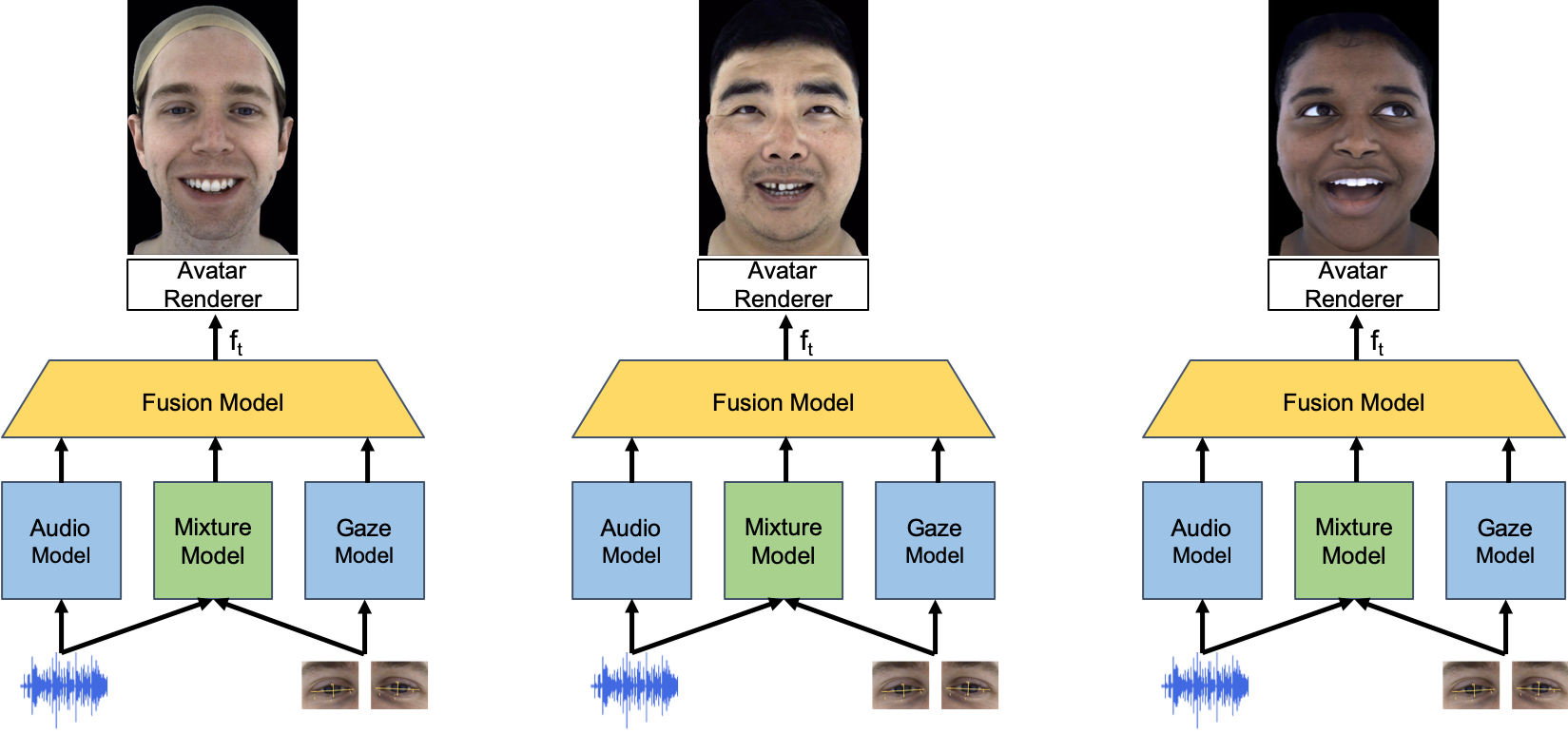}
    \caption{Our Multimodal VAE-based model predicts facial coefficients that animate a photorealistic ``Codec'' avatar model using only audio and gaze as input. Face images are renders from coefficients generated by our model.}
    \label{fig:overview}
\end{figure}

% Recent work has shown it is possible to represent photorealistic avatar faces by jointly modeling geometry and texture using data collected in a high-fidelity capture studio \cite{lombardi2018deep}. 
% It has even been shown that these avatars can be driven in a VR headset using mouth- and eye-directed cameras. However, there are no consumer-grade headsets with mouth-directed cameras on the market. 
% In this paper we describe an approach for driving photorealistic avatars using audio, which is available on practically all modern VR headsets, and eye-directed cameras, which are readily available as an aftermarket research platform.\footnote{which company did that Stanford paper get theirs from? Also, StarGaze? Star something?} 

Recent audio-driven facial animation efforts have suffered from a severe lack of data that often consist of only minutes worth of high quality facial capture (i.e.,~\cite{voca2019cvpr,nvidia2017siggraph}). 
One goal of this work was to investigate what kind of data is necessary to build an expressive audio-focused animation model. To this end we collected over 5 hours of data
%in-house
consisting of expressive, dyadic conversations across three people which were processed using the capture pipeline in~\cite{lombardi2018deep} with paired audio, gaze, and facial coefficients per-timestep. 
% While this is not practical for use cases outside of research \TODO{formulate this with care} 
This allows us to investigate the impact of training on different data subsets, understand how diverse data actually needs to be, and understand if our input modalities, and the corresponding models, are sufficient for achieving plausibly accurate \textit{expressive} facial animation. 

Multimodal fusion is a challenging problem, especially when using lossy input modalities such as audio and gaze, because there is not enough signal in either modality to accurately predict a facial expression. 
% Audio and gaze input signals are very lossy so naturally the quality of our results do not match that of camera-driven face tracking (i.e., \cite{lombardi2018deep,wei2019siggraph,chencao2019SiggraphAsia})
% or more common phone/webcam-based setups~\cite{chencao2019SiggraphAsia}. 
% Audio- and gaze-driven facial animation is inherently a more difficult problem because there is not enough signal in the audio or gaze domains alone to accurately predict a facial expression. 
While there are clear correlations between speech and lip shapes there may not be any indication of when someone smiles, raises their eyebrows, or when they open their mouth to preempt another speaker. 
Likewise, gaze direction has a clear effect on the eyes but it is unclear how gaze affects the lower face motion. 
% What this means is that in order to drive natural, expressive facial animation you must synthesize these motions. 
% In this paper we investigate approaches for multimodal fusion 
% Multimodal fusion is a deceptively challenging problem. %Standard approaches compare ideas such as early-fusion and late-fusion. 
% Recent work has described how for high-capacity models (i.e., most deep networks) it is easy to overfit to a dominant modality instead of learning complementary information~\cite{AMLpaper}. 
Standard fusion approaches tend to produce more muted facial motions where each input modality plays a more fixed role, \eg gaze only affects the eyes and audio affects the lower face.
We describe a simple-yet-effective approach for dynamically updating how the model attends to each sensor to jointly encode correlations such as how motion from eye tracking features affect expressions like smiling.
% attends to each part of the face

\fakesubsection{Contributions.}
Our main contributions are summarized as:
\begin{enumerate}
    \item[(a)] This is the first paper to provide an extensive study on the variety of expressivity data (e.g., excited conversations, descriptive tasks) required to animate natural facial motion for photorealistic avatars only from audio or from audio and eye tracking. 
    % With one exception [Nvidia] we are not aware of any papers that go in depth on the content of the data collected. 
    Other approaches rely on tiny amounts of high quality ``expressive'' data (e.g., 3-5 minutes~\cite{nvidia2017siggraph} versus our 5 hours) or moderate amounts of neutral sentence reading~\cite{cudeiro2019capture,taylor2012siggraph,taylor2016interspeech}. 
    %or large amounts of unconstrained 2D in-the-wild videos [Obama, GAN papers].
    \item[(b)] We are the first to demonstrate a real-time solution for this problem using non-linear, photorealistic full-face models of geometry and texture. Note this is harder than geometry-alone \cite{voca2019cvpr,nvidia2017siggraph} due to non-linearities in texture-based tongue motions and lip articulations.
    % We are the first to analyze different types of expressive speech data (e.g., excited, descriptive) the impact on a given model.
    % , and why this variety of data is necessary to model authentic traits of subjects that expose varying neutrality/excitement in their communication.
    \item[(c)] We discuss a fundamental issue with deep multimodal models where the network effectively learns to ignore one modality. To overcome, we describe a set of learning techniques, e.g. reconstruction of input modalities. We show quantitatively that this improves performance when paired with our dynamic, per-parameter multimodal fusion model, \cf Section~\ref{sec:model_ablations}.
    % These techniques enabled us to maintain important subtle facial cues not depicted in, e.g., [10,19].
    % we provide detailed analysis on how to effectively train the model to maintain important subtle facial cues (see Sec.\ 5.1) not depicted in, e.g., [10,19].
    % Our techniques help overcome 
    % This is an important problem that we know other researchers have also faced. 
    % Our techniques (adaptive modality weighting, reconstruction of input modalities to maintain subtle cues, etc.) are not bound to the proposed model and are likely  may help other researchers to advance their approaches. 
\end{enumerate}

\noindent
At run-time our input is synced audio and gaze and output is a vector of facial coefficients for an avatar. 
% Note that this system is real-time capable but for practical purposes all results in this paper and the supplemental video are processed offline. 
We suggest viewing the supplemental video before reading this paper.
\section{Related Work}

% \fakesubsection{Learned Audio-driven Animation.}
Recent audio-driven animation efforts approaches have focused on driving lip articulations \cite{disney2017siggraph,obama2017siggraph,visemenet2018siggraph}, full-face geometry~\cite{nvidia2017siggraph,eskimez2018iva,dave2018interspeech,snap2018icassp}, and holistic video approaches~\cite{vgg2017bmvc,zhou2018arxiv,song2018arxiv}. 
Our work focuses on geometry and texture-based full-face animation. We find encoding dynamic changes in texture is critical for realistic lip and tongue motion.
% but have made progress on subsets of this problem. 
% Our approach addresses the difficult problem of full-face animation, using a joint representation of geometry and texture, but we use metrics that are more broadly applicable to general video-based results.
% , which is strictly harder than lip-articulation, 
% Unlike many of these papers, we approach the problem of full-face animation of geometry and color. 

\fakesubsection{Lower-face Synthesis.} Taylor \etal~\cite{disney2017siggraph} and Suwajanakorn \etal~\cite{obama2017siggraph} generate lower-face animation (offline) by taking low level audio features (Phoneme-based in~\cite{disney2017siggraph} and MFCCs in~\cite{obama2017siggraph}) and predicting a set of coefficients corresponding to 2D Active Appearance Models. 
Results from Taylor \etal \cite{disney2017siggraph} are reasonable for neutral speech but lack nuanced facial motion or non-neutral expression. 
% They extracted a large set of audio features derived from phonemes, use them as input into a shallow Neural Net \TODO{check details}, and predict a set of AAM coefficients used to animate the lower face. 
% They demonstrated that their model could be retargeted to stylized characters, but did not show full-face photorealistic avatar results, which have a much higher bar for realistic animation. 
Suwajanakorn \etal~\cite{obama2017siggraph} provide compelling videos of former President Barack Obama speaking, however, upper face expression comes from reference video and is not predicted. 
% in which they composite their lower-face predictions onto reference videos of former President Barack Obama speaking, but note that any expression displayed on the upper face comes from said reference videos and is not predicted by their model. 
% This paper focused largely on an approach for compositing the predicted lower face onto another video in an offline manner. 
% Suwajanakorn \etal~\cite{obama2017siggraph} learn a PCA-based active appearance model (AAM) of the lower face that they drive using an LSTM-based audio model using MFCC features. Their model was trained on former President Obama's weekly addresses and contained substantial variation in lighting conditions, head pose, and expression. This paper focused largely on an approach for compositing the predicted lower face onto another video in an offline manner. 
Zhou \etal (VisemeNet)~\cite{visemenet2018siggraph} show how data-driven approaches -- using one hour of lower-face landmark data -- can be used to drive a set of artist-friendly visemes and jaw and lip (JALI) controls. While they improve lip articulation over their previous JALI model~\cite{jali2016siggraph}, they do not show expression such as smiles, smirks, or non-speech. 

\fakesubsection{3D Geometry.}
Karras \etal~\cite{nvidia2017siggraph} use 3-5 minutes of tracked 3D geometry, per actor, to generate expressive speech animation using linear predictive coding (LPC) audio features.
While generating full-face animation is much harder than lower-only, there is relatively poor lip closure and 
% subjective quality -- especially around the lips -- is lacking compared to other approaches likely due to a insufficient amount of data and there is 
substantial eyebrow swim.
Similarly, Cudeiro \etal~\cite{voca2019cvpr} collected around 3 minutes of speech for each of the 12 participants which they mapped to a learned FLAME~\cite{FLAME:SiggraphAsia2017} geometry model. They achieved good lip closure but because their captures all consisted of neutral speech the results are very monotone. 
%These limitations are likely due to the small amount of data with which they train their models. 
% Their predictions are more expressive than the likes of the Obama paper, the lip articulations and general motion are subjectively of low quality  \TODO{make more formal}.
Greenwood \etal~\cite{dave2018interspeech} and Eskimez \etal~\cite{eskimez2018iva} also look at full face animation but only predict sparse landmark-based marker positions which hide a lot of nuance included in high fidelity photorealistic avatars. 
A key limitation in many of these approaches is that they rely on phoneme- or phoneme-like approaches, which inherently remove stylistic cues important for expressive speech.
% As a result it is unclear if their results would generalize to photorealistic avatars. \colin{'it is unclear': raises the question why we didn't check. Can we safely say their results don't generalize to photorealistic avatars?}
% to generate full-face speech animation from a sparse set of 2D face landmarks. \TODO{..}

\fakesubsection{Image-based Animation.}
There has been recent interest in animating frontal face images using data in-the-wild (i.e., \cite{vgg2017bmvc,pantic2018arxiv,song2018arxiv,pantic2019ijcv}). Typically these GAN-based papers take a single image and generate a video as if the person is speaking. While impressive, they could not be used for our class of VR use cases which assume parametric models of the face. 
Brand \cite{brand1999} did some of the earliest work in this area, far pre-dating GANs, by computing trajectories on a manifold of possible facial motions. 
Chung \etal~\cite{vgg2017bmvc} generate full-face animation using cropped frontal images using videos in-the-wild. While their approach is inherently photorealistic, their model seemingly only animates the lower face. Recent work by Vougioukas \etal~\cite{pantic2018arxiv}, Song \etal~\cite{song2018arxiv}, and Zhou \etal~\cite{zhou2018arxiv} have used GANs to add or improve quality of full-face expression for frontal face images.
%TODO @Alex check if ArXiv paper have been published by now

% Vougioukas/Pantic: https://sites.google.com/view/facialsynthesis/home
% Zhou: https://liuziwei7.github.io/projects/TalkingFace
% Song: https://arxiv.org/pdf/1804.04786.pdf

%\TODO{remove?}Most of the work described above uses low-level audio features or phoneme-based inputs. The phoneme-based models inherently lose important information about the expression of the user. Using more expressive audio features is an interesting area where there has been a small amount of related work in speech synthesis (e.g., style embeddings ~\cite{google2018}) but is ultimately outside of the scope of this paper.

% [Snap]\cite{snap2018iccassp} had an audio-video model... 

\fakesubsection{Traditional Audio-driven Animation.}
Existing audio-driven approaches generate reasonable quality lip animations on low-fidelity stylized avatars~\cite{OculusTechNote,AmazonSumerian}. 
% \footnote{\href{https://developer.oculus.com/documentation/audiosdk/latest/concepts/book-audio-ovrlipsync/}{Oculus Lipsync}, \href{https://aws.amazon.com/sumerian/}{Amazon Sumerian}}
Solutions rely on artist-defined lip shape models (\textit{visemes}) and assume a mapping between phonemes and lip articulations.
% \footnote{\href{https://developer.oculus.com/blog/tech-note-enhancing-oculus-lipsync-with-deep-learning/}{Tech Note: Enhancing Oculus Lipsync with Deep Learning}}
Extensions to the viseme model look uncanny when applied to photorealistic avatars, in part because of their inability to distinguish between expressions (i.e., talking in an excited versus sad manner) \cite{taylor2012siggraph,taylor2016interspeech,jali2016siggraph,visemenet2018siggraph}. 
% There are two reasons why the traditional approach does not generalize to our setting:
Photorealistic avatars are frequently built on learned non-linear representations that cannot be combined with artist-created sculpts~\cite{lombardi2018deep,disney2017siggraph,nvidia2017siggraph}.
% and require more nuanced facial animation than stylized avatars to overcome the uncanny valley. 
% Our avatars require high-quality, full-face animation to look realistic. 

Other early work in this area~\cite{cao2005,cosker2003} demonstrated audio-driven animation on photorealistic avatars such as with active appearance models. Cao \etal~\cite{cao2005} shows compelling expressive animation but is based on a motion graph that requires offline post-processing to time-warp and blend motion snippets. Cosker \etal~\cite{cosker2003} also works offline by synthesizing a coherent texture map and stitching together different regions of the face.

\fakesubsection{Multimodal and Time-series Modeling.}
We build upon foundational work on variational autoencoders (VAEs)~\cite{kingma2014auto,vandenoord2017vqvae,jang2017gumbel} and temporal convolutional networks (TCNs) \cite{vandenoord2016wavenet,lea2017temporal,koltun2018arxiv}. 
% and provide baselines using modality transfer as in~\cite{gupta2016cross}.
% We define a training procedure such that regardless of the combination of input modalities the output leverages the complementary signals. 
Our approach uses the idea of learning a shared latent space among different modalities, which has been explored previously \cite{mor2019universal,liu2017unsupervised}. 
% Unlike other work learning shared, multimodal embedding~\cite{mor2019universal, liu2017unsupervised} 
Unlike \cite{mor2019universal} our approach does not need explicit regularization of the latent space by an adversarial loss and unlike \cite{liu2017unsupervised} ours learns a direct mapping from input to target modalities instead of learning a style/domain transfer.
See~\cite{morencyMultimodal} for an overview on multimodal modeling.
% Our approach is most similar to the recent preprint by Shi \etal~\cite{moe2019Neurips} on Variational, Multimodal Mixture-of-Experts Autoencoders, however, in our case we learn a set of mixture weights for our model instead of assuming each input modality should be weighed equally. 
% Recent work has described how for high-capacity models (i.e., most deep networks) it is easy to overfit to a dominant modality instead of learning complementary information~\cite{FB2019WhyMultiModalHard} \colin{edit}
Our approach attempts to overcome the following phenomenon: with high-capacity multimodal models it is easy to overfit to one modality and thereby ignore another, which has also been investigated in two recent preprints~\cite{FB2019WhyMultiModalHard,moe2019Neurips}. 
Wang \etal~\cite{FB2019WhyMultiModalHard} describe an approach that identifies when each modality starts to overfit -- using a held-out set -- and introduces gradient blending to prevent one modality from dominating the prediction. 
A preprint by Shi \etal~\cite{moe2019Neurips} describes a Mixture of Experts VAE similar to ours except they assume that each input modality should provide overlapping information: i.e., averaging modality-specific predictions provides a good estimate. In our case audio and gaze are complementary and thus we dynamically, per latent parameter, identify how each modality should be combined based on available signals at the time.
% While there has been some work on combining modalities for facial animation (e.g., audio with mobile cameras~\cite{snap2018icassp}), 

% VisemeNet uses 1 hour w/ small number of controls. This is a controlled environment so we don't have to worry about issues such as changes in lighting, visual appearance, head rotation (e.g., Obama paper; VGG@BMVC paper). 

% Work in speech synthesis, including WaveNet~\cite{wavenet} and VQ-VAE~\cite{VQVAE}, has shown that autoregressive convolutional models can be used to generated increasingly realistic speech. 

%%%%%%%%% TECHNICAL DETAILS
\section{Multimodal VAE}

% \subsection{Overview}
Our goal is to generate realistic facial motion corresponding to a photorealistic 3D avatar using only audio and gaze as input. 
We use the deep appearance model of Lombardi \etal~\cite{lombardi2018deep} and denote a sequence of facial coefficients as $ \mathbf{f} = (f_1,\dots,f_T)$ for all $T$ time steps. Each vector of coefficients $ f_t \in \mathbb{R}^{D_{\mathbf{f}}} $ is decoded into a mesh and texture map using the facial decoder proposed in~\cite{lombardi2018deep}.
% and is rendered out to video or displayed in a VR environment.
We assume corresponding audio and gaze input features $ \mathbf{a} = (a_1,\dots,a_T) $ and $ \mathbf{g} = (g_1,\dots,g_T) $ where $ a_t \in \mathbb{R}^{D_\mathbf{a}} $, $ g_t \in \mathbb{R}^{D_\mathbf{g}} $. 
Audio features, gaze, and facial coefficients are sampled at 100 Hz.
% Note that both input sequences are assumed to have the same length. 
% Given the input modalities we predict a sequence of facial coefficients $ \mathbf{f} = (f_1,\dots,f_T), f_t \in \mathbb{R}^{D_{\mathbf{f}}}$. 
% During training, we have corresponding facial coefficients for each audio/gaze pair, \ie we are given a set of temporal sequence triplets $ (\mathbf{a}, \mathbf{g}, \mathbf{f}) $.

% In this work we introduce a multimodal model that captures the temporal dynamics of input modalities and output coefficients. 
% features using temporal convolutional networks as encoders and decoders.
% We generate a video of our 3D avatar using the predicted coefficients $ \mathbf{f} $ using the face renderer proposed in \cite{lombardi2018deep}.
A straightforward approach, and one that we use as a baseline, is to train a TCN-based regressor that takes in a sequence of audio and gaze features and simply predicts the facial coefficients for each time-step. This is similar to what is done in~\cite{voca2019cvpr} but applied to both of our modalities. 
We show that this approach does not handle nuanced and complementary interactions between each input modality, such as correlations between eye gaze and smiles or speech and blinking.
We propose an alternative approach using a specially structured VAE that learns a shared mapping across sensor types and facial configurations, as shown in Figure~\ref{fig:model}.
This model has modality-specific encoders, modality-specific decoders, a facial coefficient decoder, and a mixture encoder that determines how to combine information from each modality. 
%We show that our approach is better able to leverage correlations across input modalities. 
At training time, we force the model to not only predict facial coefficients from audio and gaze input but also to reconstruct both input modalities.
This strategy forces the model to focus on eyes and mouth movement and improves the fidelity of the avatar.
At test-time we only need the encoders and facial coefficient decoder, however, we find that reconstructing the input modalities at training time improves our model's ability to generalize to unseen data. 
% we only reconstruct the facial coefficients from which the 3D avatar is rendered.

% Given the input modalities we predict a sequence of facial coefficients $ \mathbf{f} = (f_1,\dots,f_T), f_t \in \mathbb{R}^{D_{\mathbf{f}}}$. During training, we have corresponding facial coefficients for each audio/gaze pair, \ie we are given a set of temporal sequence triplets $ (\mathbf{a}, \mathbf{g}, \mathbf{f}) $.
% Our model captures the temporal dynamics of input and output features using temporal convolutional networks as encoders and decoders.
% We generate a video of our 3D avatar using the predicted coefficients $ \mathbf{f} $ using the face renderer proposed in \cite{lombardi2018deep}.
% Given the generated facial coefficients as output of the face decoder, the 3D avatar is then rendered for each from from $ \mathbf{f} $ using the face renderer proposed in \cite{lombardi2018deep}.

\subsection{The Model}
\label{sec:model}

\begin{figure}[t]
    \centering
    \includegraphics[scale=0.7]{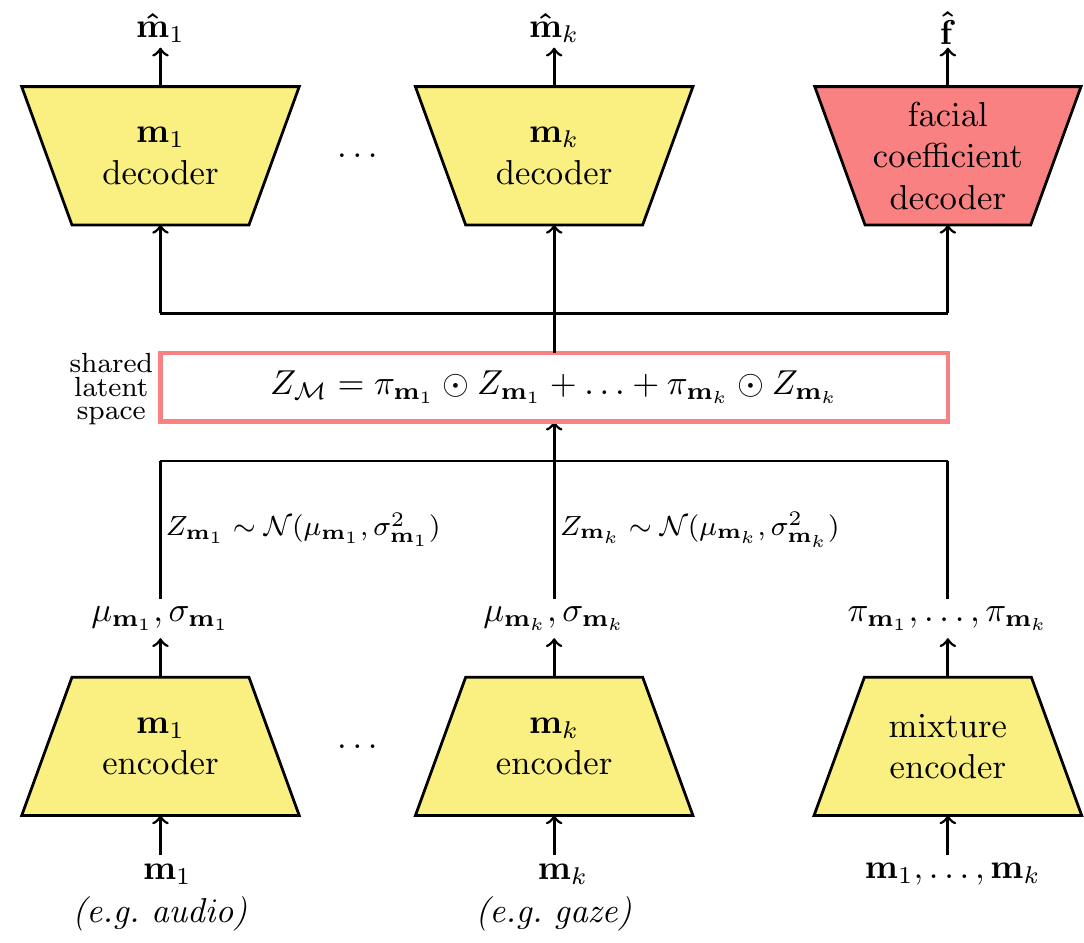}
    \caption{Our Multimodal VAE architecture. Given $k$ input modalities $ \mathbf{m}_1,\dots,\mathbf{m}_k$ ($k=2$ for audio and gaze in our case), $k$ encoders compute an embedding $ Z_{\mathbf{m}_i} $ for each modality. A mixture encoder outputs a weight for each modality which is then used to compute the shared latent embedding $ Z_\mathcal{M} $. A set of decoders reconstruct all input modalities and facial coefficients $ \mathbf{f} $ from $ Z_\mathcal{M} $.}
    \label{fig:model}
\end{figure}

We start by setting notation and describing the standard VAE~\cite{kingma2014auto} and then extend it to our multimodal VAE.

\fakesubsection{Preliminaries.} Consider a VAE that maps input $ \mathbf{x} $ to a latent space $Z_\mathbf{x}$,
\ie the VAE learns an encoder $ q(Z_\mathbf{x}|\mathbf{x}) $ and decoder $ p(\mathbf{x}|Z_\mathbf{x}) $ that maximize the evidence lower bound,
\begin{align}
    \mathrm{ELBO}_\mathbf{x} = \mathbb{E}_{Z_\mathbf{x}} [\log p(\mathbf{x}|Z_\mathbf{x})] - \mathrm{KL}[q(Z_\mathbf{x}|\mathbf{x}) || p(Z_\mathbf{x})]
\end{align}
with $ \mathrm{KL}[\cdot] $ denoting the Kullback-Leibler divergence.
As in~\cite{kingma2014auto}, we assume the latent prior $ p(Z_\mathbf{x}) $ is an isotropic Gaussian with unit variance and the encoder $ q(Z_\mathbf{x}|\mathbf{x}) $ models a Gaussian distribution with mean $ \mu_\mathbf{x} $ and diagonal covariances $ \sigma_\mathbf{x}^2 $.
Optimizing the decoder $ p $ with the $ \ell_2 $-loss, the maximization of the ELBO is equivalent to minimizing the loss
\begin{align}
    \mathcal{L}_\mathbf{x} = \|\mathbf{x} - \mathbf{\hat x}\|^2 + \mathrm{KL}[q(Z_\mathbf{x}|\mathbf{x}) || p(Z_\mathbf{x})],
    \label{eq:l2face}
\end{align}
where $ \mathbf{\hat x} $ is the input reconstructed from the latent embedding $ Z_\mathbf{x} $, \ie the output of the decoder $ p $. The $ \mathrm{KL} $ term can be seen as a regularizer on the latent space, pushing it towards an isotropic Gaussian. Note that optimizing the reconstruction error using the $ \ell_2 $ loss $ \|\mathbf{x} - \mathbf{\hat x}\|^2 $ corresponds to maximizing $ p(\mathbf{x}|Z_\mathbf{x}) $, assuming the distribution to be an isotropic Gaussian with mean $ \mathbf{\hat x} $.

% While this formulation encodes audio, gaze, and/or facial coefficients into a Gaussian latent space, and can reconstruct each, it cannot be used to take the sensor inputs and predict facial coefficients.

\fakesubsection{Multimodal VAE.}
\label{sec:technical:mulmodVAE}
We formulate an alternative VAE architecture, depicted in Figure~\ref{fig:model}, that encodes multiple input modalities $ \mathbf{m}_i \in \mathcal{M} $ ($i = 1,\dots,k) $ into a shared latent space $ Z_\mathcal{M} $ using a mixture of per-modality embeddings.
The decoder should be able to reconstruct all input modalities from this shared latent space, such that the model is forced to maintain sufficiently detailed information about the input modalities in the shared latent space.
% Extending the loss from Equation~\eqref{eq:l2face} accordingly, 
The multimodal loss is then $ \mathcal{L} = \mathcal{L}_\mathrm{rec} + \mathcal{L}_\mathrm{KL} $, where
\begin{align}
    \mathcal{L}_\mathrm{rec} &= \sum_{\mathbf{m} \in \mathcal{M}} \|\mathbf{m} - \mathbf{\hat m}\|^2, \label{eq:rec} \\
    \mathcal{L}_\mathrm{KL} &= \mathrm{KL}\big[q(Z_\mathcal{M}|\mathbf{m}_1,\dots,\mathbf{m}_M \in \mathcal{M}) || p(Z_\mathcal{M})\big]. \label{eq:kl_loss}
\end{align}
The latent embedding $ Z_\mathcal{M} $ depends on all input modalities and is regularized towards an isotropic Gaussian prior $ p(Z_\mathcal{M}) $. 

Our input modalities are audio features $ \mathbf{a} $ and gaze features $ \mathbf{g} $, \ie $ \mathcal{M} = \{\mathbf{a}, \mathbf{g}\} $. 
At run-time our goal is to predict a set of facial coefficients, so we add an additional decoder that is optimized to predict the facial coefficients from the joint audio and gaze embedding $ Z_\mathcal{M} $, see Figure~\ref{fig:model} for an illustration.
Formally, the reconstruction loss from Equation~\eqref{eq:rec} then becomes
\begin{align}
    \mathcal{L}_\mathrm{rec} = \|\mathbf{f} - \mathbf{\hat f}\|^2 + 
    \sum_{\mathbf{m} \in \mathcal{M}} \|\mathbf{m} - \mathbf{\hat m}\|^2,
\end{align}
\ie all input modalities and the facial coefficients are reconstructed from the shared latent embedding $ Z_\mathcal{M} $.

\fakesubsection{Joint Embedding of Audio and Gaze.} % ($ Z_\mathcal{M} $)
\label{sec:technical:jointEmbedding}
Our fusion approach, illustrated in Figure~\ref{fig:model},
% In the following, we describe our fusion approach
takes an arbitrary set of $ M $ modalities $ \mathbf{m}_i \in \mathcal{M} $ using 
% , $ M $ separate encoders first 
predicted mean and standard deviation $ \mu_{\mathbf{m}} $ and $ \sigma_{\mathbf{m}} $ for each input modality $ \mathbf{m} $ such that
\begin{align}
    Z_\mathbf{m} \sim \mathcal{N}(\mu_\mathbf{m}, \sigma_\mathbf{m}^2),
    \label{eq:normal_modality}
\end{align}
where $ \mu_\mathbf{m}, \sigma_\mathbf{m}^2 \in \mathbb{R}^L $ with $ L $ being the dimensionality of the latent space.
A separate ``mixture" encoder predicts mixing coefficients $ \pi_\mathbf{m} $ for each modality.
% $ \pi_\mathbf{m}  \in \mathbb{R}^{D_\mathbf{m}} $
% $ \mathbf{m} \in \mathcal{M} $
% .
 $ \pi_\mathbf{m} $ is a vector containing a mixture weight for each component of the latent space 
% \footnote{For simplicity of notation, we treat the multiplication of two vectors as element-wise operations.}
% The weights $ \pi_\mathbf{m} $
and is generated using a softmax layer so each element is positive and sums to one across modalities: $ \sum_\mathbf{m} \pi_{\mathbf{m},d} = 1$ for each coefficient $ d $.
The shared embedding $ Z_{\mathcal{M}} $ is defined as a weighted sum of the latent random variables $ Z_\mathbf{m} $ of each individual modality,
\begin{align}
    Z_{\mathcal{M}} = \sum_{\mathbf{m} \in \mathcal{M}} \pi_\mathbf{m} \odot Z_\mathbf{m},
\end{align}
where $ \odot $ is the Hadamard product. Then,
\begin{align}
    Z_{\mathcal{M}} \sim \mathcal{N}\Big(\sum_{\mathbf{m} \in \mathcal{M}}\pi_\mathbf{m} \odot \mu_\mathbf{m}, \sum_{\mathbf{m} \in \mathcal{M}}\pi_\mathbf{m}^2 \odot \sigma_\mathbf{m}^2\Big).
    \label{eq:normal_shared}
\end{align}
During training, we regularize the joint embedding $ Z_\mathcal{M} $ to follow an isotropic Gaussian prior using the KL divergence.

% While we described the overall model architecture in the previous section, how to model the joint latent space $ Z_\mathcal{M} $ still needs to be determined.
% ($ \mathcal{M} = \{\mathbf{a}, \mathbf{g}\} $)
This approach %for learning a multimodal embedding $ Z_\mathcal{M} $  
was designed with two properties in mind. 
First, each modality should be disentangled within the latent space such that, if necessary, each modality can simultaneously drive a different part of the face.
For example, if a user speaks while darting their eyes around, the audio component should help determine the mouth shapes and the eye gaze should determine the eye shapes and upper face expressions. 
This suggests that the mixing weights should be defined on a per-coefficient basis such that audio can drive speech-related mouth shapes while eye gaze simultaneously drives eye and upper face shapes.
Second, at each point in time the model should be able to dynamically identify which modality is more useful for animating certain facial expressions.
When a user is talking then the audio should drive the lip shapes, however, if there is silence then correlations with eye gaze may help indicate other facial expressions, such as smiling.
Similarly, if there is a substantial amount of background noise in the audio the model should learn to ignore this signal without explicitly affecting the gaze signal. 
We achieve this by introducing the a weighting function that outputs the weights $ \pi_\mathbf{m} $, updated at each time-step, to determine the importance of each modality for each latent coefficient.

\fakesubsection{Implications of the model formulation.}
In contrast to a conventional regression model, this multimodal VAE has several advantages.
First, reconstructing the original input modalities along with the facial coefficients forces the model to maintain detailed information about both audio and gaze in the shared latent embedding.
We find this results in %This allows for higher fidelity and 
more accurate lip closure and eye movement of the avatar.
Second, the shared and weighted latent space provides some interpretability.
More precisely, the model can explicitly decide on the importance of each modality depending on the temporal context.
Third, the VAE formulation with the Kullback-Leibler loss on the latent space is more robust against noise and improves the overall quality of the model.
An empirical evaluation in Section~\ref{sec:model_ablations} shows the improvements of our proposed multimodal VAE over a conventional regression baseline and ours without the Kullback-Leibler loss.

\subsection{Network Architecture}

Our encoders and decoders are simple temporal convolutional networks (TCNs) \cite{lea2017temporal} with skip connections. Each branch consists of one $1 \times 1$ convolution to resize the dimensionality of the input (per-frame) to a fixed number of channels (e.g., $c=128$), a stack of temporal convolutions,\footnote{We use $5$ layers per stack with kernel length=$5$ frames and dilation=$2^l$ for layer $l$.}
and a set of output $1 \times 1$ convolutions for generating $\mu$ and $\sigma$ (for encoders) or a single output for each decoder. 
\textit{Leaky ReLU} activations with leakage coefficient $ 0.2 $ are used after each convolution and skip connections are used between every set of stacked convolutions. 
We investigated other architectures, including using mechanisms such as key-value attention \cite{vaswani2017attention}, but ultimately found that results were only incrementally better than this simpler TCN architecture. 
%The model has a receptive field of $ 2.5 $ seconds. Experiments are run using acausal convolutions, meaning, a real-time system would have a $ 1.25 $ second latency.

%%%%%%%%% EXPERIMENTS
\section{Experimental Setup}

\fakesubsection{Data.}
We captured 5 hours of high-density 3D scans of three subjects from different ethnic backgrounds and gender using a multi-camera capture system.
Figure~\ref{fig:overview} shows images of all three subjects.
Tracked 3D meshes, a tracked deep active appearance model, and head pose were extracted in a similar manner as described in Lombardi \etal~\cite{lombardi2018deep}. 
The subjects performed different types of tasks during the capture that comprised a wide variety of facial expressions and social interactions, \eg reading sentences, describing images, summarizing videos, trivia games, and conversations with another person.
For perspective, \cite{nvidia2017siggraph} and \cite{voca2019cvpr} used 3 - 5 minutes of 3D data per person. % and~\cite{voca2019cvpr} use less than one hour of data.
A frame-level deep face model was trained following~\cite{lombardi2018deep} on a $ 45 $ minute subset of the data for each subject. Using the resulting appearance encoder, each frame is then mapped to a $ 256 $-dimensional vector of facial coefficients.
Later, the avatar is rendered using this appearance model which generates texture and geometry from our predicted facial coefficients, \cf Figure~\ref{fig:overview}.
Two tasks are held out per subject for evaluation which corresponds to roughly $ 25 $ minutes of test data each. 
The remaining data is used for training.
The evaluation sequences are (a) a conversational task containing a variety of natural expressions such as laughter and smiles and (b) an image description task with mostly neutral facial expression and a stronger focus on lip synchronization.

\begin{figure}[t]
    \centering
    \includegraphics[scale=0.17]{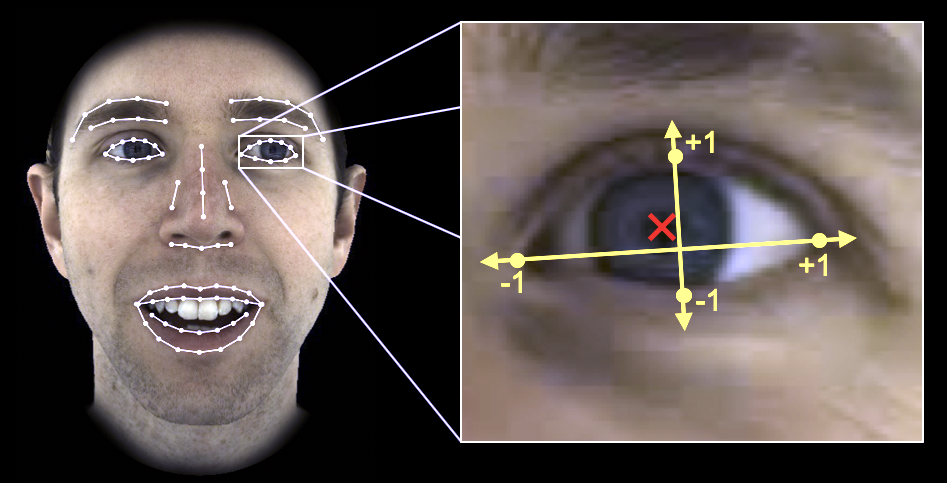}
    \caption{Left: facial landmarks used for evaluation. Right: gaze features are pupil coordinates in a normalized coordinate system defined by the eye corners and iris diameter.}
    \label{fig:landmarks}
\end{figure}

\fakesubsection{Audio Features.}
We extract $ 80 $-dimensional mel spectrograms from the raw $ 16 $kHz wave signal using the \textit{torchaudio} pyTorch package.
This feature extractor computes a spectrogram using a short-time Fourier transform (STFT) every $ 10 $ms over a $ 50 $ms Hanning window and warps the resulting $ 2,048 $ frequency bins at each timeframe onto an $ 80 $-dimensional feature vector using the Mel-scale. 
Mel spectrograms are widely used in tasks such as speech recognition~\cite{chiu2018state} and audio-visual speech processing~\cite{afouras2018deep}.

%Many existing audio-driven approaches use phoneme-based features due to their ability to generalize across people and to be robust across microphones and environments (i.e.,~\cite{disney2017siggraph,obama2017siggraph}). Despite being robust, phonemes tend to abstract away important signal necessary for encoding expression. We compare our mel spectrogram results with a model using phoneme probabilities as input using a model based on~\cite{OculusTechNote}.
% \footnote{\href{https://developer.oculus.com/blog/tech-note-enhancing-oculus-lipsync-with-deep-learning/}{Oculus LipSync Tech Note}}
% which shares similarity to a Time-Delay Neural Net trained with Kaldi~\cite{Kaldi}.

\fakesubsection{Gaze Features.}
We compute 2D gaze coordinates for each pupil using a normalized coordinate system as shown in Figure~\ref{fig:landmarks} (right).
The left and right corners of the eye are defined as coordinates $ (-1, 0) $ and $ (1, 0) $, and the perpendicular axis is scaled using the radius of the iris.
This representation is invariant to scaling, translation, and rotation within the image plane.
The pupil coordinates were extracted from a frontal view of each user but conceptually this information could also be extracted from a gaze tracker in a VR headset.
Accurate upper face expression (i.e., eyebrow motion) likely requires training models directly on raw images from eye-directed cameras in a VR headset. This was investigated by Wei \etal~\cite{wei2019siggraph} using custom multi-camera hardware. The data collection and processing pipelines necessary are highly non-trivial and are beyond our scope.

\fakesubsection{Evaluation Metrics.}
Subtle facial cues are difficult to quantify but have a huge influence on human perception. 
While we find the metrics described below to be informative, ultimately we recommend viewing the video in the supplemental material for subjective evaluation.

For quantitative evaluation we render both the ground truth 3D avatar reconstructions and the generated avatars as videos with a resolution of $ 960 \times 640 $ and measure errors after running a commercial facial landmark tracker (see Figure~\ref{fig:landmarks}). We report the mean squared error on the landmarks between the original data and the generated avatars for different facial regions, \ie (1) for the 32 landmarks on the mouth, (2) for the 13 landmarks on the nose, (3) for the 20 landmarks around the eyes, (4) for the 18 landmarks on the eyebrows, and (5) averaged over all facial landmarks.

Lip closure is especially important for perceptual quality and is assessed by a second metric. 
We detect all lip closures in the recorded data by determining when the landmarks on the inner upper and lower lip match
(\ie when they do not deviate by more than two pixels each)
and compare them to the lip closures detected in the generated avatars. 
% High precision and high recall on lip closure so 
We report the F1-score to emphasize the importance of both high precision and high recall.

%We therefore measure the distance between rendered faces and recorded faces at several facial landmarks around the eyes, eyebrows, nose, and mouth. Note that particularly mouth and eyes capture highly relevant information about lip synchronization and gaze that strongly impact the authenticity of rendered faces. Since lip closure that fits the audio signal has a particularly high impact perceptually, we also consider this for our evaluation.

%We run a commercial facial landmark tracker on both the recorded ground truth and the rendered faces.\footnote{All faces are rendered with a resolution of $ 960 \times 640 $.} The extracted landmarks are shown in Figure~\ref{fig:landmarks}. The mean squared pixel error is then reported for different facial regions, \ie (1) for the 32 landmarks on the mouth, (2) for the 13 landmarks on the nose, (3) for the 20 landmarks around the eyes, and (4) for the 18 landmarks on the eyebrows, respectively. Additionally, the mean squared pixel error over all facial landmarks is reported.

\section{Experiments \& Analysis}
\label{sec:experiments}

We describe results for three ablation studies: a comparison of various models, investigations on training data configurations, and the impact of different input modalities.

% ---------------------------------------------------------------------------
\subsection{Model Ablations}
\label{sec:model_ablations}

\begin{table*}[t]
    \tablesmall
    \centering
    \caption{a-d: effect of the KL loss on different latent embeddings. d vs.\ e: effect of reconstructing audio and gaze features during training vs.\ only predicting facial coefficients alone. f: a conventional regression baseline.}
    \begin{tabularx}{0.8\textwidth}{lXrrrrrrr}
        \toprule
        &                                               & \multicolumn{5}{c}{Landmark Error ($\downarrow$)}                             & & F1-score ($\uparrow$) \\
        &                                               & eyebrows   & eyes       & nose       & mouth      & all        & & \textit{lip closure}            \\
        \midrule
        %\multicolumn{8}{l}{\textit{Effect of the KL loss}} \\
        (a) & no KL loss                                    &  $ 4.27 $ & $ 6.84 $ &  $ 2.27 $ & $ 20.46 $ & $ 15.15 $ & &  $ 0.557 $   \\
        (b) & KL on $ Z_\mathbf{m} $ ($\mathbf{m} \in \{\mathbf{a},\mathbf{g}\}$) &  $ 4.18 $ & $ 6.08 $ &  $ 2.12 $ & $ 19.03 $ & $ 14.12 $ & &  $ \mathbf{0.569} $   \\
        (c) & KL on $ Z_{\mathcal{M}} $           &  $ \mathbf{4.00} $ & $ 5.62 $ &  $ \mathbf{1.99} $ & $ \mathbf{17.95} $ & $ \mathbf{13.36} $ & &  $ 0.546 $   \\
        (d) & KL on $ Z_\mathbf{m} $ ($\mathbf{m} \in \{\mathbf{a},\mathbf{g}\}$) and $ Z_{\mathcal{M}} $           &  $ 4.06 $ & $ \mathbf{5.49} $ &  $ 2.06 $ & $ 18.42 $ & $ 13.42 $ & &  $ 0.521 $   \\
        \cmidrule(lr){1-9}
        (e) & no audio/gaze reconstruction &  $ 4.18 $ & $ 6.05 $ &  $ 2.15 $ & $ 19.42 $ & $ 14.22 $ & &  $ 0.534 $   \\
        (f) & conventional regression      &  $ 4.38 $ & $ 7.43 $ &  $ 2.34 $ & $ 20.64 $ & $ 15.52 $ & &  $ 0.462 $   \\
        \bottomrule
    \end{tabularx}
    \label{tab:ablationStudy}
\end{table*}

\begin{figure*}[t]
    \centering
    \begin{minipage}{0.58\textwidth}
        \hspace{0.5cm}
        \includegraphics[scale=0.7]{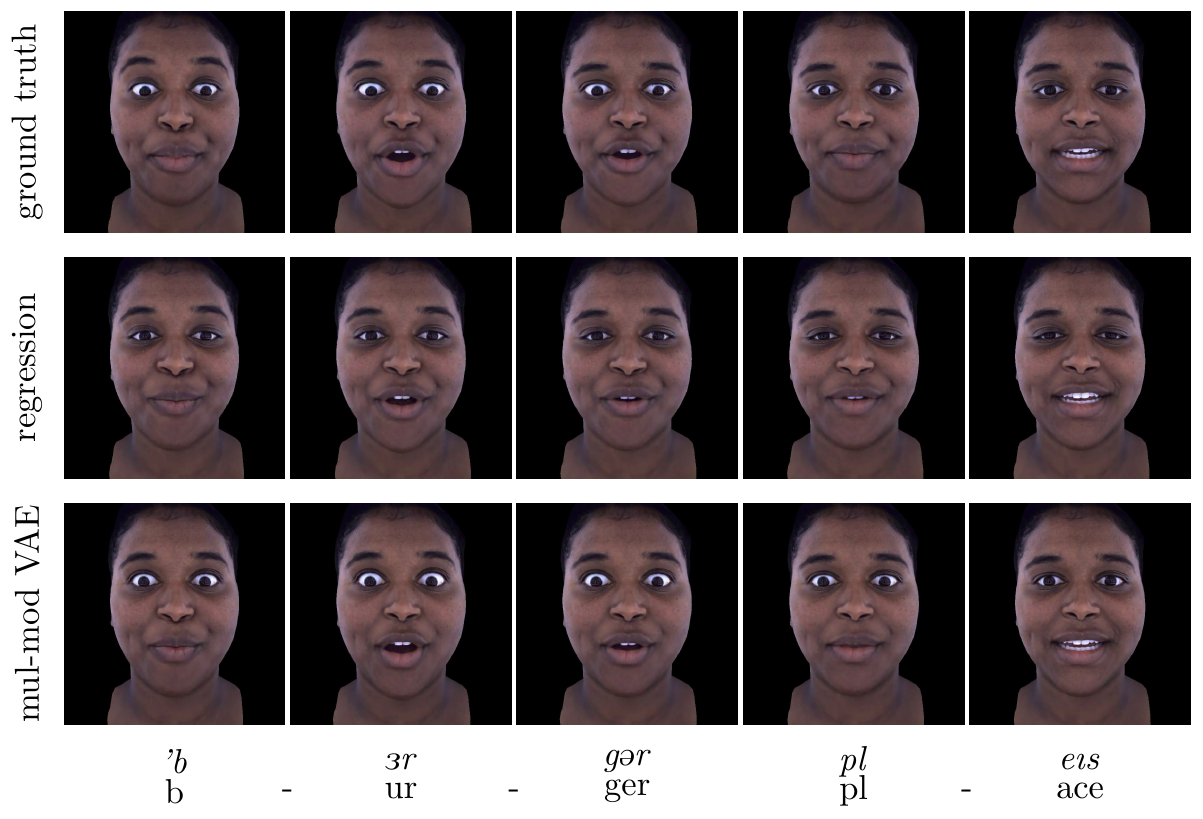}
        
        \tablesmall (a) True, regression, and Multimodal VAE results for utterance \textit{``burger place.''} The regression baseline fails to close the lips at \textit{pl} and does not capture the wide-opened eyes or raised brows.
    \end{minipage}
    \hfill
    \begin{minipage}{0.35\textwidth}
        \includegraphics[scale=0.7]{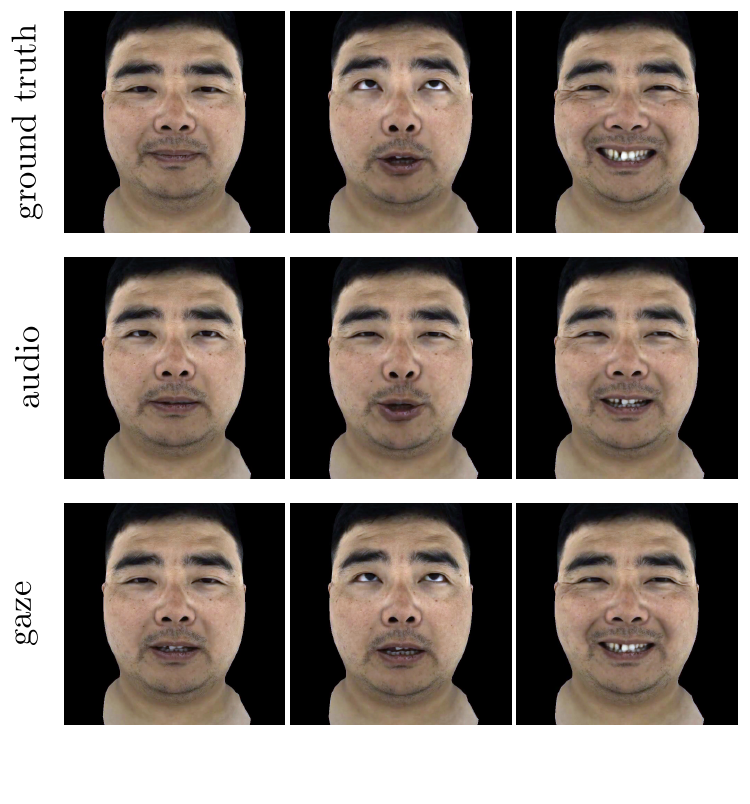}
        
        \tablesmall (b) Comparison of gaze-only and audio-only models which each encode complementary signals. \phantom{lalala lalala lalala lalala} %Note how gaze picks up voiceless expressions like smiles
    \end{minipage}
    \caption{Qualitative results of the rendered avatar.}
    \label{fig:qualitative}
\end{figure*}

In this section, we analyze the components of our model and compare the multimodal VAE to a conventional TCN-based regression network. We show (a) that a structured shared latent space leads to better models, (b) that reconstructing the input modalities is beneficial for subtle facial motion such as accurate lip closure, and (c) that our proposed multimodal VAE yields more authentic and expressive facial motion than a conventional regression baseline.

\begin{table*}[t]
    \tablesmall
    \centering
    \caption{Comparison of audio and gaze models on tasks that are neutral (image description) and expressive (conversation). Note the impact of gaze on estimating the mouth shape and the impact of audio on accurate lip closure.}
    \begin{tabularx}{0.8\textwidth}{lXrrrrrrr}
        \toprule
                                               &                & \multicolumn{5}{c}{Landmark Error ($\downarrow$)}                             & & F1-score ($\uparrow$) \\
                                               &                & eyebrows   & eyes       & nose       & mouth      & all        & & \textit{lip closure}            \\
        \midrule
        \multirow{3}{*}{\textit{conversation}} & only audio     & $ 5.03 $ & $ 13.70 $ &  $ 4.19 $ & $ 32.67 $ & $ 25.70 $ & &  $ 0.437 $   \\
                                               & only gaze      & $ 4.20 $ &  $ \mathbf{5.92} $ &  $ 3.11 $ & $ 32.94 $ & $ 20.03 $ & &  $ 0.062 $   \\ 
                                               & audio + gaze   & $ \textbf{3.66} $ &  $ 6.05 $ &  $ \textbf{2.49} $ & $ \textbf{22.95} $ & $ \textbf{15.71} $ & &  $ \textbf{0.450} $   \\
        \cmidrule(lr){1-9}
        \multirow{3}{*}{\textit{descriptive}}  & only audio     & $ 5.05 $ & $ 9.88 $ &  $ 1.73 $ & $ 13.08 $ & $ 14.05 $ & &  $ 0.650 $   \\
                                               & only gaze      &  $ 4.78 $ &  $ 5.27 $ &  $ 1.89 $ & $ 22.95 $ & $ 14.88 $ & &  $ 0.075 $   \\
                                               & audio + gaze   &  $ \textbf{4.49} $ &  $ \textbf{5.01} $ &  $ \textbf{1.30} $ & $ \textbf{10.79} $ & $ \textbf{9.99} $ & &  $ \textbf{0.677} $   \\
        \bottomrule
    \end{tabularx}
    \label{tab:audioVsGaze}
\end{table*}

\fakesubsection{Structuring the Shared Latent Space.} 
Conventional VAEs learn a structured latent space by imposing a Gaussian prior on the latent embeddings.
For our multimodal VAE, there are several strategies where to apply this prior.
The KL loss can either be applied to the per-modality embeddings $ Z_\mathbf{m} $, to the joint embedding $ Z_\mathcal{M} $ as proposed in Section~\ref{sec:technical:jointEmbedding}, or to all $ Z_\mathbf{m} $ \textit{and} $ Z_\mathcal{M} $.
A comparison of the landmark errors and lip closure scores of the corresponding experiments in Table~\ref{tab:ablationStudy}b-d shows that structuring only the per-modality embeddings is beneficial for lip closure but degrades the overall facial geometry and appearance.
Imposing the KL loss on the shared embedding $ Z_\mathcal{M} $, on the contrary, leads to consistent improvements in the landmark error in all facial regions at the cost of only a minor degradation in lip closure.

Applying the KL loss on the per-modality embeddings and the shared embedding (Table~\ref{tab:ablationStudy}d) is inferior to regularizing $ Z_\mathcal{M} $ only.
Investigation of Equation~\eqref{eq:normal_modality} and Equation~\eqref{eq:normal_shared} shows that -- in case the per-modality and shared embeddings are both regularized towards an isotropic Gaussian -- the KL loss is minimized if the mixure weights are either zero or one.
Therefore, regularizing both, individual and shared embeddings, strips the model of the ability to interpolate between the embeddings of different modalities.

Training the multimodal VAE without a KL loss leads to an unstructured latent space that is not particularly suitable for interpolation between the observed training embeddings and therefore does not generalize well enough on unseen data.
We empirically observe this issue in Table~\ref{tab:ablationStudy}a, where the landmark error is consistently worse than for any of the KL regularized variants in lines b-d.

\fakesubsection{Reconstructing the Input Modalities.}
A comparison of line c and e in Table~\ref{tab:ablationStudy} reveals that reconstructing the input modalities improves performance. If the shared latent space does not contain sufficient information about gaze and audio, the model tends to overfit the training data and fails to generate accurate lip closure, mouth shape, or eye movement.
We find the reconstruction of the input modalities particularly important to generate subtle and fine-grained facial motion and expressions.

\fakesubsection{Multimodal VAE vs.\ Regression Baseline.}
% A natural baseline for the task we address in this work is a conventional regression from audio and gaze features to facial coefficients.
% Therefore, 
We trained a baseline regression-based TCN that receives the concatenated audio and gaze features as input and predicts the facial coefficients directly. The size of the TCN is comparable to our multimodal VAEs. 
Conceptually this is similar to what Cudeiro \etal~\cite{voca2019cvpr} used for audio-alone except in our case we use audio and gaze.
We find the baseline performs significantly worse than the proposed multimodal VAE in landmark errors and lip closure (Table~\ref{tab:ablationStudy}c and f).
The regression model tends towards more neutral expressions and fails to capture more articulated expressions such as excitement with widely opened eyes and raised eyebrows or subtle lip motion and accurate lip closure for plosives such as ``\textit{p}" (Figure~\ref{fig:qualitative}a).

\begin{figure}[t]
    \centering
    \includegraphics[scale=0.15]{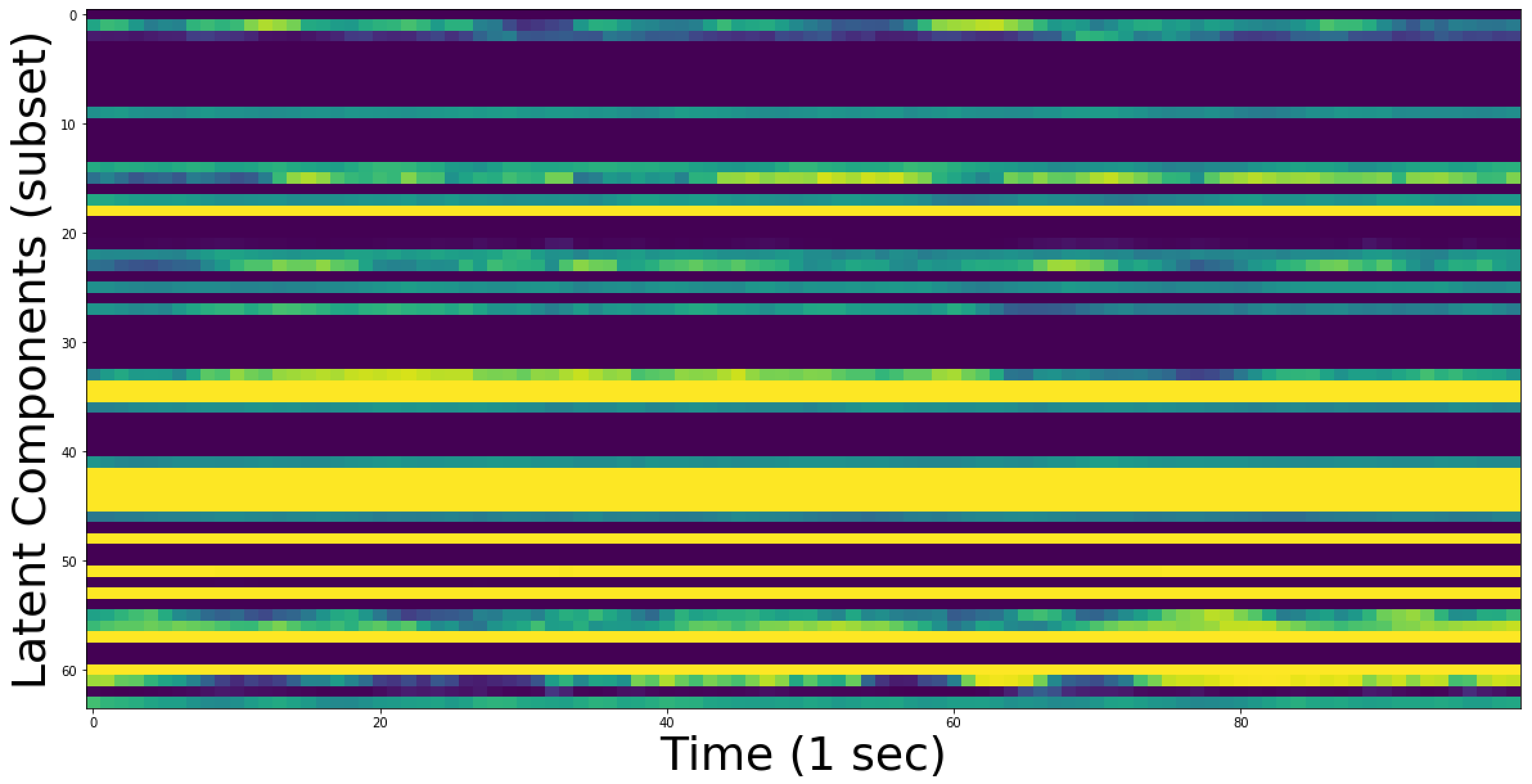}
    \caption{Mixture weights (per parameter) from a 1 second test clip. Some latent components (yellow rows) always use the audio embedding, others (blue rows) always use the gaze embedding. The rest (varying colors) change dynamically based on the input.}
    \label{fig:weights}
\end{figure}

One strength of the multimodal VAE is its capability to dynamically decide on the weight each modality gets.
Figure~\ref{fig:weights} shows the mixture weights of the audio modality for each component of the latent space over a one second long snippet.
Interestingly, the model learns to use some latent components exclusively for a single modality.
Specifically, the blue horizontal stripes represent components for which the audio weight is always zero, \ie components that exclusively encode gaze information.
The yellow horizontal stripes are purely dedicated to the audio modality.
Many latent components use a mixture of the input modalities that varies over time depending on the temporal context.

% ---------------------------------------------------------------------------
\subsection{Modality Ablations}

\begin{figure}[t]
    \centering
    \includegraphics[scale=0.38]{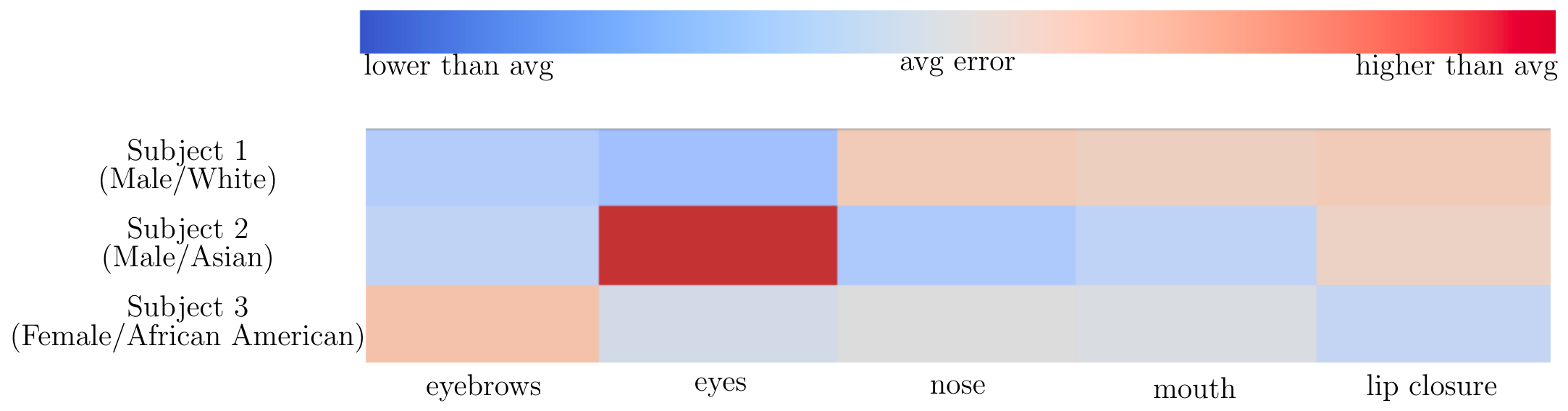}
    \caption{Relative errors per subject compared to the errors averaged over all subjects. Red indicates errors that are proportionally larger than the averaged errors, blue is lower than average.}
    \label{fig:subject_comparison}
    \vspace{-0.3cm}
\end{figure}

\fakesubsection{Subject Comparisons.}

We compare our full system with single modality versions to analyze the impact each signal has on encoding different types of facial expressions. 
Experiments were evaluated on ``image description" and ``conversational" tasks with a single-input version of the architecture outlined in Figure~\ref{fig:model}, \ie without the weights encoder and decoders for the held-out modalities.
% For training, we rely on  with the difference that the audio/gaze encoder operates on a single modality only, so there is no need for the weights network and the second encoder. 
% Moreover, during training only the used input modality -- audio or gaze -- is reconstructed (see red part of decoder box in Figure~\ref{fig:model}).
% We do not reconstruct the held-out modality in the decoder. 
See results in Table~\ref{tab:audioVsGaze}. 
As expected, the landmark error around the eyes is large when only using audio and low when using the gaze input.
It was also of no surprise that the gaze modality fails to predict lip closure.
Performance for the mouth shape estimation was more surprising.
For the image description task, the audio modality accurately models the mouth shape estimation and the gaze modality fails to achieve comparably good results.
%We found that facial expression during descriptive speech is usually relatively neutral and the the range of motion around the mouth is relatively small. 
%Thus, the model can achieve low error by simply learning a good average mouth shape. 
For the conversational task, surprisingly both modalities have huge errors for the mouth shape.
In this task the user frequently smiles, laughs, and gives the other person quizzical expressions.
While the audio may pick up on speech related lip motion and loud laughter, we find that non-verbal, voiceless expressions like smiling are more reliably predicted from gaze, see Figure~\ref{fig:qualitative}b for an example.
Within the given data captures, smiles and laughter have strong co-occurring gaze patterns where the user squints their eyes and looking downwards, and can therefore be picked up without audio input.
While each modality is indeed complementary, combining both results substantially improved performance as shown in Table~\ref{tab:audioVsGaze}.

% ---------------------------------------------------------------------------
\subsection{Data Ablations}

It should be no surprise that the kind of data used for training is critical for the results.
In this section, we investigate the impact of different subjects, monotone versus expressive tasks, and different audio features.

Figure~\ref{fig:subject_comparison} shows a per-subject breakdown of the errors relative to the averaged errors over all subjects.
Red means the error is proportionally larger than for the average over subjects, blue means it is lower.
The largest outlier is the eye error for Subject 2 
% , a male Asian person.
% Compared to the other subjects,
who has has smaller eyes than the other subjects and a dark iris which makes pupil tracking less reliable.
It is also interesting to observe results from Subject 3, a trained actress, in the supplementary video.
She heavily moves her eyebrows during expressive speech, which is hard to synthesize correctly from only audio and gaze and leads to increased errors in that facial region.
We attribute Subject 1's slightly increased mouth landmark errors and lip closure errors to his frequent open-mouthed smiles and his tendency during conversations to open his mouth while silent to show an intent to speak.
These voiceless mouth movements particularly impact lip closure and mouth shape prediction.

\begin{table}[t]
    \tablesmall
    \centering
    \caption{Impact of training data on different test tasks for landmark error and lip closure (Subject 1).}
    \begin{tabularx}{0.45\textwidth}{X|cc}
        \toprule
        \textbf{train data}          & \multicolumn{2}{c}{\textbf{test data}} \\
                            & descriptive \#2                    & conversation \#2 \\
        \midrule
                            & \textit{$\downarrow$ landmarks / lips $\uparrow$} & \textit{$\downarrow$ landmarks / lips $\uparrow$}  \\
        descriptive \#1         & $ \phantom{1}8.38\ /\ 0.715 $           & $ 35.66\ /\ 0.234 $ \\
        conversation \#1        & $ 13.22\ /\ 0.429 $           & $ 14.56\ /\ 0.216 $ \\
        %\midrule
        all                 & $  \phantom{1}8.97\ /\ 0.712 $           & $ 16.06\ /\ 0.275 $ \\
        \bottomrule
    \end{tabularx}
    \label{tab:training_data}
\end{table}

\fakesubsection{Training Data Characteristics.}
%The choice of training data is critical for generating plausible and accurate facial motion.
While deep neural networks excel at interpolating within the training distribution, they typically struggle to extrapolate beyond it.
This is what has limited most prior work to generating monotone-looking speech; typical datasets simply include neutral read sentences as their training data.
We show that a diverse dataset with both expressive and descriptive speech is key to not only accurate but also authentic facial animation.

%\colin{rambling paragraph. rewrite/remove}
%The most important thing to consider when using data-driven approaches for audio-driven facial animation is that the source of your data is critical for generating plausibly accurate facial motion. This is what has limited most prior work to generating monotone-looking speech; typical datasets simply include neutral read sentences as their training data. Audio and/or gaze are very lossy and are insufficient for generating physically accurate 
%predictions of facial motion. Regardless of your input data or machine learning model, you are going to place biases on the types of expressions that you are able to generate using purely learned approaches. 

Table~\ref{tab:training_data} shows the performance of our approach when training the same model on three different subsets of the data. The \textit{descriptive} tasks tend to be largely monotone, the \textit{conversational} tasks tend to be more lively, and \textit{all} includes a mixture of many types of expressive and descriptive speech.
We evaluate on held out descriptive speech and expressive conversation to illustrate how the nature of training data affects different test scenarios.
As expected, when training on the \textit{descriptive} tasks and testing on \textit{conversational} tasks the landmark error is poor, because the model does not generate as many expressions like smiling and laughter. Likewise when trained on \textit{conversational} tasks the \textit{descriptive} results deteriorate since all generated facial motion is more extreme than it should be. 
Overall, we find that descriptive data is responsible for accurate lip closure whereas conversational data is crucial to capture a wide range of expressions.
As one would hope, when training on both, descriptive \textit{and} expressive speech, the model generates both muted and expressive results when needed.
% With both kinds of data in the training set, the results are most convincing.

\begin{figure}[t]
    \centering
    \includegraphics[scale=.6]{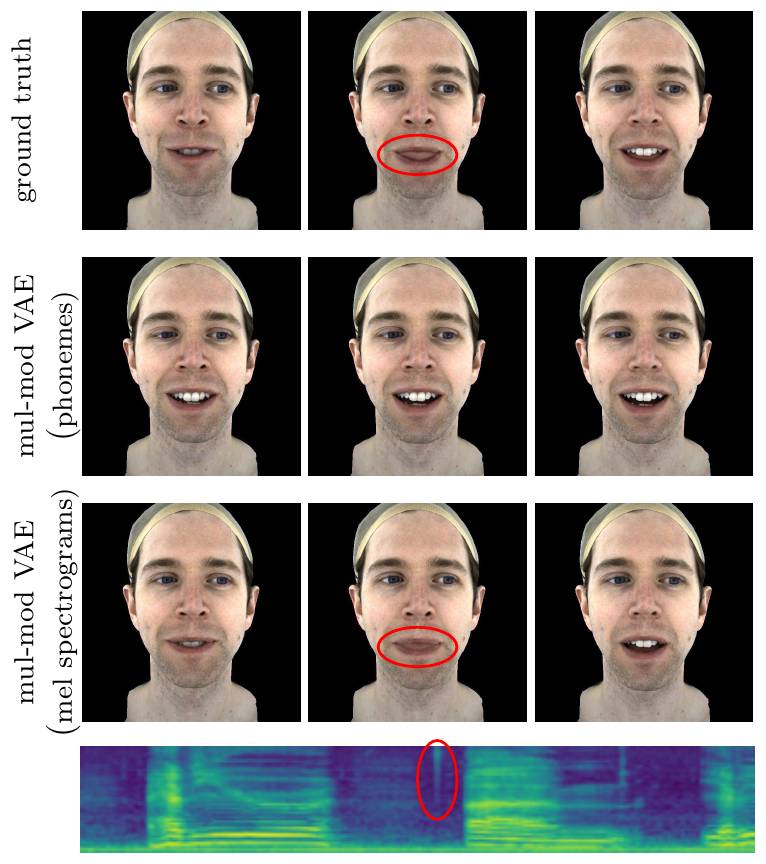}
    \caption{Our model encodes subtle audio cues such as wetting your lips with your tongue, which are only represented by a short high-frequency segment in the spectrogram.}
    \label{fig:tongue}
    \vspace{-0.5cm}
\end{figure}

\fakesubsection{Phonemes vs.\ Mel Spectrograms.}
Phonemes are a popular mid-level representation in speech processing~\cite{bisani2008joint,graves2013speech} and also find application in face animation~\cite{disney2017siggraph,jali2016siggraph,visemenet2018siggraph}.
We therefore compare results when using phonemes versus mel spectrograms.
Phoneme models are trained to robustly recognize which out of 43 phonemes a user is saying and inherently abstract away the subtleties of speech. Figure~\ref{fig:tongue} shows that we are capable of picking up very nuanced sounds such as licking your lips which is encoded by a tiny blip in the high frequency information. This is not encoded in the phoneme-based model. While this example may not generalize to low quality microphones or noisy environments, it highlights one of many signals that are lost from phoneme-based encodings. 
% For example, you can say the same thing with many expressions and the resulting phoneme predictions will be the same. 
We find that the kind of training data, as described above, has a larger impact than the choice of audio features, however, Table~\ref{tab:phones_vs_melspec} indicates a substantial improvement in lip closure using mel spectrograms and a modest improvement in expressivity. This is shown in more detail in the supplemental video.

\begin{table}[t]
    \tablesmall
    \centering
    \caption{Impact of different audio features (Subject 1).}
    \begin{tabularx}{0.45\textwidth}{Xrrr}
        \toprule
                            & Mouth Error ($\downarrow$) & & lip closure ($\uparrow$) \\ %(F1 score) ($\uparrow$) \\
        audio features      & (landmark error)           & & (F1 score) \\
        \midrule
        phonemes            & $ 20.74 $                     & & $ 0.251 $ \\
        mel spectrograms    & $ 20.53 $                     & & $ 0.465 $ \\
        \bottomrule
    \end{tabularx}
    \label{tab:phones_vs_melspec}
\end{table}

\section{Conclusion}

% \fakesubsection{Limitations.}
% While the results of our approach are visually appealing and reflect convincing facial motion, the method has two important limitations.
% First, mel spectrograms are sensitive to changes in the acoustic environment.
% For our experiments, the audio data has been recorded in a quiet environment.
% It is not trivial to transfer this approach \textit{into the wild}, \ie into noisy environments.
% Second, both the audio modality and the photorealistic avatar are person-specific representations. Scaling both to a vast amount of possibly thousands of subjects without the requirement of more than just a few minutes of data per person is challenging and requires further work that goes beyond the scope of this paper.

In this work we showed that with a sufficient amount of expressive animation data we are able to map from raw audio to expressive facial animation. 
We find that in general our approach to multi-modal fusion is able to overcome limitations with models overfitting to individual sensors and improves animation performance.
In the supplemental video we show that the quality of lip articulations from our audio-only solution can even surpass the quality from video-based solutions such as~\cite{wei2019siggraph} which uses mouth and eye cameras. 
Future work may look at combining our audio-driven model with this video-based solution. 
% We have started to address this problem of combining our audio-driven model with this video-based solution. 
% In our early investigations in this area, we have combining camera and audio 
% from~\cite{wei2019siggraph} with our audio-based system. 
% In our early investigations we have encountered the same types of sensor fusion limitations highlighted in this paper, and were able to overcome them using the same principles as introduced here.
% While in this paper we only discussed sensor fusion between audio and eye tracking, in followup work fusing multiple video streams, we have identified the same type of sensor fusion limitations as described here 

\fakesubsection{Ethics Remarks.}
This work, and work on photorealistic avatars more broadly, have strong implications on privacy
and should be approached with caution if considering real-world use cases. Restricting avatar access, for example with biometrics, is critical for preventing misrepresentation. Furthermore users should be made aware of how their avatar is being portrayed, potentially in real-time, to prevent issues where the predicted facial expression does not match the user's intent.

% ---------------------------------------------------------------------------
% ---------------------------------------------------------------------------
% ---------------------------------------------------------------------------

%\fakesubsection{Limitations.}
%There are several limitations that must be overcome to generate more natural and expressive photorealistic facial animation for VR and AR. 
%Some of the largest are as follows. The phoneme probabilities used here as audio features abstract away most information about a user's expression. 
%While these performed better in our informal experiments than spectral features or learned audio embeddings, there is significant room for improvement. 
%In this work, we used gaze as a proxy for eye cameras that could be placed in a VR headset. 
%In theory, using actual camera images of the eye as input instead of relying on gaze predictions may improve the ability to detect facial expressions. 
%Although our model does a reasonable job predicting facial expression given our lossy sensors, we find the amount of variation in expressions is relatively limited (i.e., only few types of 'laughter' and 'smile' encoded). 
%This could be improved through investigation into newer stochastic temporal model (e.g., Flow-based generative models~\cite{WaveGlow}). 
%Lastly, while we mitigated lip closure issues within our model design, it may be advantageous to jointly learn a speech model with the deep facial representation.

{\small
\bibliographystyle{ieee_fullname}
\bibliography{references}

\begin{thebibliography}{10}\itemsep=-1pt

\bibitem{AmazonSumerian}
Amazon sumerian, 2018.
\newblock \url{https://aws.amazon.com/sumerian/}. Last accessed 2 Aug 2020.

\bibitem{afouras2018deep}
Triantafyllos Afouras, Joon~Son Chung, Andrew Senior, Oriol Vinyals, and Andrew
  Zisserman.
\newblock Deep audio-visual speech recognition.
\newblock {\em IEEE Transactions on Pattern Analysis and Machine Intelligence},
  2018.

\bibitem{koltun2018arxiv}
Shaojie Bai, J.~Zico Kolter, and Vladlen Koltun.
\newblock An empirical evaluation of generic convolutional and recurrent
  networks for sequence modeling.
\newblock {\em arXiv:1803.01271}, 2018.

\bibitem{morencyMultimodal}
Tadas Baltrusaitis, Chaitanya Ahuja, and Louis{-}Philippe Morency.
\newblock Multimodal machine learning: {A} survey and taxonomy.
\newblock {\em IEEE Transactions on Pattern Analysis and Machine Intelligence},
  41(2):423--443, 2019.

\bibitem{MagicLeap}
James Bancroft, Nafees~Bin Zafar, Sean Comer, Takashi Kuribayashi, Jonathan
  Litt, and Thomas Miller.
\newblock Mica: a photoreal character for spatial computing.
\newblock In {\em ACM SIGGRAPH 2019 Talks}, 2019.

\bibitem{bisani2008joint}
Maximilian Bisani and Hermann Ney.
\newblock Joint-sequence models for grapheme-to-phoneme conversion.
\newblock {\em Speech communication}, 50(5):434--451, 2008.

\bibitem{brand1999}
Matthew Brand.
\newblock Voice puppetry.
\newblock In {\em ACM Transaction on Graphics}, 1999.

\bibitem{snap2018icassp}
Xin Chen, Chen Cao, Zehao Xue, and Wei Chu.
\newblock Joint audio-video driven facial animation.
\newblock In {\em IEEE Int. Conf. on Acoustics, Speech and Signal Processing},
  pages 3046--3050, 2018.

\bibitem{chiu2018state}
Chung-Cheng Chiu, Tara~N. Sainath, Yonghui Wu, Rohit Prabhavalkar, Patrick
  Nguyen, Zhifeng Chen, Anjuli Kannan, Ron~J. Weiss, Kanishka Rao, Ekaterina
  Gonina, et~al.
\newblock State-of-the-art speech recognition with sequence-to-sequence models.
\newblock In {\em IEEE Int. Conf. on Acoustics, Speech and Signal Processing},
  pages 4774--4778, 2018.

\bibitem{vgg2017bmvc}
Joon~Son Chung, Amir Jamaludin, and Andrew Zisserman.
\newblock You said that?
\newblock In {\em British Machine Vision Conference}, 2017.

\bibitem{cosker2003}
Darren Cosker, David Marshall, Paul~L. Rosin, and Yulia Hicks.
\newblock Video realistic talking heads using hierarchical non-linear
  speech-appearance model.
\newblock In {\em MIRAGE}, 2003.

\bibitem{voca2019cvpr}
Daniel Cudeiro, Timo Bolkart, Cassidy Laidlaw, Anurag Ranjan, and Michael~J.
  Black.
\newblock Capture, learning, and synthesis of {3D} speaking styles.
\newblock {\em IEEE Conf. on Computer Vision and Pattern Recognition}, 2019.

\bibitem{cudeiro2019capture}
Daniel Cudeiro, Timo Bolkart, Cassidy Laidlaw, Anurag Ranjan, and Michael~J.
  Black.
\newblock Capture, learning, and synthesis of 3d speaking styles.
\newblock In {\em IEEE Conf. on Computer Vision and Pattern Recognition}, 2019.

\bibitem{jali2016siggraph}
Pif Edwards, Chris Landreth, Eugene Fiume, and Karan Singh.
\newblock {JALI}: An animator-centric viseme model for expressive lip
  synchronization.
\newblock {\em ACM Transaction on Graphics}, 35(4), 2016.

\bibitem{siren2018gdc}
{Epic Games}, 3Lateral, Tencent, and Vicon.
\newblock {Siren: World's first digital human}.
\newblock In {\em Game Developers Conference}, 2018.

\bibitem{eskimez2018iva}
Sefik~Emre Eskimez, Ross~K. Maddox, Chenliang Xu, and Zhiyao Duan.
\newblock Generating talking face landmarks from speech.
\newblock In {\em Int. Conf. on Latent Variable Analysis and Signal
  Separation}, pages 372--381, 2018.

\bibitem{google2017headsetRemoval}
Christian Frueh, Avneesh Sud, and Vivek Kwatra.
\newblock Headset removal for virtual and mixed reality.
\newblock In {\em ACM SIGGRAPH 2017 Talks}, SIGGRAPH '17, pages 80:1--80:2, New
  York, NY, USA, 2017. ACM.

\bibitem{graves2013speech}
Alex Graves, Abdel-rahman Mohamed, and Geoffrey Hinton.
\newblock Speech recognition with deep recurrent neural networks.
\newblock In {\em IEEE Int. Conf. on Acoustics, Speech and Signal Processing},
  pages 6645--6649, 2013.

\bibitem{dave2018interspeech}
David Greenwook, Iain Matthews, and Stephen~D. Laycock.
\newblock Joint learning of facial expression and head pose from speech.
\newblock In {\em Interspeech}, pages 2484--2488, 2018.

\bibitem{jang2017gumbel}
Eric Jang, Shixiang Gu, and Ben Poole.
\newblock Categorical reparameterization with gumbel-softmax.
\newblock In {\em Int. Conf. on Learning Representations}, 2017.

\bibitem{OculusTechNote}
Sam Johnson, Colin Lea, and Ronit Kassis.
\newblock Tech note: Enhancing oculus lipsync with deep learning, 2018.
\newblock
  \url{developer.oculus.com/blog/tech-note-enhancing-oculus-lipsync-with\
  -deep-learning/}. Last accessed 2 Aug 2020.

\bibitem{nvidia2017siggraph}
Tero Karras, Timo Aila, Samuli Laine, Antti Herva, and Jaakko Lehtinen.
\newblock Audio-driven facial animation by joint end-to-end learning of pose
  and emotion.
\newblock {\em ACM Transaction on Graphics}, 36(4), 2017.

\bibitem{kingma2014auto}
Diederik~P. Kingma and Max Welling.
\newblock Auto-encoding variational bayes.
\newblock In {\em Int. Conf. on Learning Representations}, 2014.

\bibitem{lea2017temporal}
Colin Lea, Michael~D. Flynn, Rene Vidal, Austin Reiter, and Gregory~D. Hager.
\newblock Temporal convolutional networks for action segmentation and
  detection.
\newblock In {\em IEEE Conf. on Computer Vision and Pattern Recognition}, 2017.

\bibitem{FLAME:SiggraphAsia2017}
Tianye Li, Timo Bolkart, Michael.~J. Black, Hao Li, and Javier Romero.
\newblock Learning a model of facial shape and expression from {4D} scans.
\newblock {\em ACM Transaction on Graphics}, 36(6), 2017.

\bibitem{liu2017unsupervised}
Ming-Yu Liu, Thomas Breuel, and Jan Kautz.
\newblock Unsupervised image-to-image translation networks.
\newblock In {\em Advances in Neural Information Processing Systems}, pages
  700--708, 2017.

\bibitem{lombardi2018deep}
Stephen Lombardi, Jason Saragih, Tomas Simon, and Yaser Sheikh.
\newblock Deep appearance models for face rendering.
\newblock {\em ACM Transaction on Graphics}, 37(4), 2018.

\bibitem{lombdardi2019voxel}
Stephen Lombardi, Tomas Simon, Jason Saragih, Gabriel Schwartz, Andreas
  Lehrmann, and Yaser Sheikh.
\newblock Neural volumes: Learning dynamic renderable volumes from images.
\newblock {\em ACM Transaction on Graphics}, 38(4), 2019.

\bibitem{mor2019universal}
Noam Mor, Lior Wolf, Adam Polyak, and Yaniv Taigman.
\newblock A universal music translation network.
\newblock {\em Int. Conf. on Learning Representations}, 2019.

\bibitem{meetmike2017siggraph}
Mike Seymour, Chris Evans, and Kim Libreri.
\newblock Meet mike: Epic avatars.
\newblock {\em ACM SIGGRAPH 2017 VR Village}, 2017.

\bibitem{moe2019Neurips}
Yuge Shi, Narayanaswamy Siddharth, Brooks Paige, and Philip Torr.
\newblock Variational mixture-of-experts autoencoders for multi-modal deep
  generative models.
\newblock In {\em Advances in Neural Information Processing Systems}, 2019.

\bibitem{song2018arxiv}
Yang Song, Jingwen Zhu, Xiaolong Wang, and Hairong Qi.
\newblock Talking face generation by conditional recurrent adversarial network.
\newblock In {\em Int. Joint Conf. on Artificial Intelligence}, 2019.

\bibitem{obama2017siggraph}
Supasorn Suwajanakorn, Steven~M. Seitz, and Ira Kemelmacher-Shlizerman.
\newblock Synthesizing obama: Learning lip sync from audio.
\newblock {\em ACM Transaction on Graphics}, 36(4), 2017.

\bibitem{taylor2016interspeech}
Sarah Taylor, Akihiro Kato, Iain~A. Matthews, and Ben~P. Milner.
\newblock Audio-to-visual speech conversion using deep neural networks.
\newblock In {\em Interspeech}, pages 1482--1486, 2016.

\bibitem{disney2017siggraph}
Sarah Taylor, Taehwan Kim, Yisong Yue, Moshe Mahler, James Krahe,
  Anastasio~Garcia Rodriguez, Jessica Hodgins, and Iain Matthews.
\newblock A deep learning approach for generalized speech animation.
\newblock {\em ACM Transaction on Graphics}, 36(4), 2017.

\bibitem{taylor2012siggraph}
Sarah~L. Taylor, Moshe Mahler, Barry-John Theobald, and Iain Matthews.
\newblock Dynamic units of visual speech.
\newblock In {\em Proc. of the ACM SIGGRAPH/Eurographics Symposium on Computer
  Animation}, pages 275--284, 2012.

\bibitem{vandenoord2016wavenet}
A{\"a}ron Van Den~Oord, Sander Dieleman, Heiga Zen, Karen Simonyan, Oriol
  Vinyals, Alex Graves, Nal Kalchbrenner, Andrew~W. Senior, and Koray
  Kavukcuoglu.
\newblock Wavenet: A generative model for raw audio.
\newblock In {\em ISCA Speech Synthesis Workshop}, page 125, 2016.

\bibitem{vandenoord2017vqvae}
Aaron van~den Oord, Oriol Vinyals, and Koray Kavukcuoglu.
\newblock Neural discrete representation learning.
\newblock In {\em Advances in Neural Information Processing Systems}, pages
  6306--6315, 2017.

\bibitem{vaswani2017attention}
Ashish Vaswani, Noam Shazeer, Niki Parmar, Jakob Uszkoreit, Llion Jones,
  Aidan~N. Gomez, {\L}ukasz Kaiser, and Illia Polosukhin.
\newblock Attention is all you need.
\newblock In {\em Advances in Neural Information Processing Systems}, pages
  5998--6008, 2017.

\bibitem{cao2005}
Konstantinos Vougioukas, Stavros Petridis, and Maja Pantic.
\newblock Expressive speech-driven facial animation.
\newblock In {\em ACM Transaction on Graphics}, 2005.

\bibitem{pantic2018arxiv}
Konstantinos Vougioukas, Stavros Petridis, and Maja Pantic.
\newblock End-to-end speech-driven facial animation with temporal gans.
\newblock In {\em British Machine Vision Conference}, 2018.

\bibitem{pantic2019ijcv}
Konstantinos Vougioukas, Stavros Petridis, and Maja Pantic.
\newblock Realistic speech-driven facial animation with gans.
\newblock In {\em International Journal on Computer Vision}, 2019.

\bibitem{FB2019WhyMultiModalHard}
Weiyao Wang, Du Tran, and Matt Feiszli.
\newblock What makes training multi-modal networks hard?
\newblock {\em arXiv:1905.12681}, 2019.

\bibitem{wei2019siggraph}
Shih-En Wei, jason Saragih, Tomas Simon, Adam~W. Harley, Stephen Lombardi,
  Michal Purdoch, Alexander Hypes, Dawei Wang, Hernan Badino, and Yaser Sheikh.
\newblock {VR} facial animation via multiview image translation.
\newblock {\em ACM Transaction on Graphics}, 38(4), 2019.

\bibitem{zhou2018arxiv}
Hang Zhou, Yu Liu, Liu Liu, Ping Luo, and Xiaogang Wang.
\newblock Talking face generation by adversarially disentangled audio-visual
  representation.
\newblock In {\em AAAI Conf. on Artificial Intelligence}, 2019.

\bibitem{visemenet2018siggraph}
Yang Zhou, Zhan Xu, Chris Landreth, Evangelos Kalogerakis, Subhransu Maji, and
  Karan Singh.
\newblock Visemenet: Audio-driven animator-centric speech animation.
\newblock {\em ACM Transaction on Graphics}, 37(4), 2018.

\end{thebibliography}
}

\end{document}